# Grammar Engineering Meets LLMs: Development of Cantonese and Irish ParGram Treebanks

**Chit-Fung Lam**
University of Oxford
University of Manchester
lawrence.lam@ling-phil.ox.ac.uk

**Elaine Uí Dhonnchadha**
Trinity College Dublin
uidhonne@tcd.ie

## Abstract

Grammar engineering requires expertise in linguistic formalism and computational implementation, especially in parallel grammar projects that balance cross-linguistic consistency with language-specific properties. This paper presents the development of Cantonese and Irish treebanks within the Parallel Grammar (ParGram) Project, where linguistic parallelism is maintained at an abstract functional level. We also investigate the methodological potential and limitations of using multilingual LLMs to support grammar engineering, focusing on Cantonese–Irish translation and the generation of formal syntactic structures using OpenAI's `gpt-oss-120b` model. The results show that translation performance was generally unsatisfactory and unaffected by prompt language. For syntactic structure generation, the model produced some structurally meaningful outputs, but performed poorly on tasks requiring cross-linguistic abstraction. Nonetheless, LLM-generated outputs may still offer some reference value by suggesting alternative analyses and (partially) capturing predicate–argument relations. Overall, our findings highlight both the potential and limitations of using LLMs in collaborative grammar engineering, while underscoring the continued importance of expert-driven analysis and verification.

## 1 Introduction

Grammar engineering involves the design and computational implementation of linguistically well-motivated grammars (Duchier and Parmentier, 2015). It requires precise, rule-based systems—often within formal frameworks such as Lexical Functional Grammar (LFG) and Head-driven Phrase Structure Grammar (HPSG)—that map natural language utterances to deep representations encoding syntactic and semantic relations (Bender, 2008; Forst and King, 2023; Zamaraeva et al., 2022). The Parallel Grammar (ParGram) Project is a large-scale, ongoing effort aimed at developing parallel LFG grammars for typologically diverse languages (Sulger et al., 2013; Butt et al., 1999). ParGram grammars are implemented in the Xerox Linguistic Environment (XLE; Crouch et al. 2011). Parsing with these grammars yields LFG's c-structure and f-structure representations (Section 2.2). T3he c-structure captures constituency, word order, and part-of-speech information, while the f-structure encodes grammatical functions (e.g., SUBJ, OBJ) and associated features such as tense, aspect, number, and gender (Dalrymple et al., 2019). ParGramBank comprises of the family of ParGram treebanks obtained by applying LFG/XLE grammars to a parallel corpus translated across languages (Rosén, 2023). These treebanks maintain a level of parallelism in the f-structure, achieved via a shared inventory of grammatical features (King, 2004), consistent with LFG's aim of capturing language universals at an abstract functional level (Kaplan and Bresnan, 1982; Bresnan et al., 2016).

Cantonese and Irish are under-resourced languages, which are typologically different from each other and most well-documented languages (McCrae and Doyle, 2019; Xiang et al., 2024). Their contrasting grammatical properties provide a challenging test case for maintaining grammar parallelism within ParGram while encoding language-specific differences. In Section 2, we describe some of the challenges encountered in representing language-specific features while maintaining grammar parallelism across languages.

Grammar engineering is inherently knowledge- and labour-intensive, requiring expertise in both linguistic theory and computational implementation. With the advent of large language models (LLMs), there has been growing interest in whether such models can assist grammar development (Spencer and Kongborrirak, 2025) and whether they exhibit the metalinguistic capabil-ities needed to produce deep linguistic analyses, including formal syntactic structures (Begu et al., 2025; Murphy et al., 2025; Kennedy, 2025). Within the ParGram framework—particularly in the construction of ParGramBank—grammar engineers face, among others, two common tasks: (i) producing reliable translations across typologically diverse languages to build parallel corpora; (ii) specifying appropriate syntactic representations for the translated sentences. These tasks are crucial for evaluating whether existing grammars can successfully parse the data and for identifying where new constraints or refinements are required. Given that LLMs may serve as assistants in translation and structural analysis, this paper investigates the extent to which they can support grammar engineering. Specifically, we evaluate whether a multilingual LLM can provide useful cross-linguistic translations and LFG-style analyses for Cantonese and Irish in the ParGramBank setting. Our goal is not to evaluate the linguistic competence of LLMs in general, but to assess their practical usefulness as tools that may assist grammar engineers in parallel grammar development.

This paper is structured as follows. Section 2 describes the Cantonese and Irish ParGram treebanks, including their design, progress, and challenges, alongside some selected linguistic phenomena. Section 3 presents the aims, methodology, and results of evaluating a multilingual LLM (OpenAI's `gpt-oss-120b`) with respect to its potential to support grammar engineering tasks using Cantonese and Irish ParGramBank data.

## 2 Cantonese and Irish ParGramBank

This section presents the Cantonese and Irish ParGram treebanks, outlining key design challenges, illustrated with selected examples from the first 50 sentences of ParGramBank. These sentences are included in Appendix A.

### 2.1 ParGram Treebank Development

ParGram treebanks are pure parsebanks in that the c-structures and f-structures they contain are generated automatically by parsing sentences with their respective LFG/XLE grammars (Rosén, 2023). No ad hoc post-parsing corrections or manual annotations are permitted. As such, ParGram treebanks differ from many traditional treebanks, which often incorporate varying degrees of manual correction or post hoc adjustment (Rosén, 2023). A key advantage of this design is that the treebank remains fully synchronized with its underlying grammar: since all c- and f-structures are produced by the same grammar, inconsistencies between analyses are avoided. However, this tight coupling also entails a trade-off, as any error in a c- or f-structure cannot be corrected directly in the treebank but instead requires revision of the grammar itself, a process that can be time-consuming. To support ongoing development and facilitate cross-linguistic consistency, the ParGram community has established a dedicated forum for discussion. Since 2023, monthly ParGram f-structure comparison meetings bring together LFG/XLE grammar engineers to address theoretical and implementation issues.[1] Grammar development is iterative: errors are typically identified through parsing failures, unexpected analyses, and comparison across ParGram languages. Regression testing using testsuite sentences is an intrinsic part of the XLE grammar development cycle.[2]

ParGramBank originally comprised parallel treebanks for twelve languages from six families, based on a shared corpus of 101 sentences. The corpus, initially derived from a tractor manual in English, French, and German, was expanded to include missing core syntactic constructions (Butt et al., 1999). The first 50 sentences, which are reported on in this paper, cover a wide range of phenomena, including transitivity, unaccusative–unergative distinctions, clause types, voice alternations, dative and double object constructions, complementation, control and raising, relative clauses, negation, phrasal verbs, prepositional phrases, copula constructions, reflexives, reciprocals, modification, expletives, existentials, and comparatives. Following Lehmann et al. (1997) and Bender et al. (2011), ParGram treats broad grammatical coverage as a central goal of grammar engineering (Sulger et al., 2013).

### 2.2 Cantonese and Irish: c-structure variation and f-structure parallelism

Cantonese and Irish are typologically distinct: Cantonese, a Sino-Tibetan language, is predominantly SVO with little morphological inflection, whereas Irish, an Indo-European language (Celtic branch), is VSO with comparatively rich inflectional morphology. Examples (1) and (2) give translation equivalents of *The driver starts the tractor* (ParGramBank, sentence 1). As shown in Figures 1 and 2, although the c-structures differ in reflecting surface order, the f-structures encode the same predicate–argument relations, specifically the verbal subcategorization for SUBJ and OBJ.[3] This demonstrates how parallelism can be maintained at the f-structural level.

[1] `https://sites.google.com/view/pargram-meetings`

[2] `https://ling.sprachwiss.uni-konstanz.de/pages/xle/doc/xle.html#TS.regression`

(1) 個 司機 啓動 拖拉機
CL driver start tractor
'The driver starts the tractor.' (Cantonese)

(2) Tosaíonn an tiománaí an tarracóir
starts the driver the tractor
'The driver starts the tractor.' (Irish)

Besides maintaining cross-linguistic parallelism, the f-structures in Figure 2 also encode language-specific differences between Cantonese and Irish in their syntax. For example, Irish exhibits tense inflection, whereas Cantonese lacks overt tense marking but encodes aspectual distinctions, including progressive, perfective, and experiential aspects (Matthews and Yip, 2013). Following King's (2004) ParGram feature conventions, the Cantonese treebank employs the feature TNS-ASP as a cross-linguistically shared feature. However, its values do not include tense specification, reflecting the properties of Cantonese. Ongoing work expands the feature–value inventory for aspect marking (Matthews and Yip, 2013). A similar contrast arises in nominal domains: Irish noun phrases are morphologically case-marked, whereas Cantonese noun phrases are not. In addition, Cantonese is a classifier language, where classifiers and measure words occur in complementary distribution within noun phrases. This is captured by the CLASS-MEASURE feature and its associated TYPE value in the Cantonese treebank, while no corresponding feature is required for Irish.

### 2.3 Design, progress, and challenges of Cantonese and Irish ParGramBank

Both the Cantonese and Irish ParGram treebanks are under active expansion, alongside the ongoing development of their LFG/XLE grammars. This paper primarily targets the first 50 sentences in ParGramBank (Section 2.1). A recurring challenge in their co-development is maintaining cross-linguistic parallelism, particularly in complex constructions where word order differences (Section 2.2) become more prominent. Examples (3) and (4) illustrate an object relative clause embedded within a copula structure (ParGramBank, sentence 19).[4] In Cantonese, the relative clause appears pre-nominal with a relativizer, whereas in Irish it is post-nominal, and differences in SV vs. VS word order arise both in the copula construction and within the embedded clause. Despite these surface differences, the resulting f-structures (Figure 3) exhibit a high degree of parallelism, achieved by representing (i) the copular complement as a PREDLINK function predicating over the SUBJ (Butt et al., 1999), (ii) the relative clause as an embedded ADJUNCT within the SUBJ, and (iii) the object gap as an empty pronoun (*null-pro*) present in f-structure but absent from c-structure. These examples show how ParGram engineers maintain f-structural parallelism while accommodating language-specific properties, and highlight the importance of incorporating linguistic insights, such as gap-based relative clause strategies (Tallerman, 2020, ch 8), without compromising computational robustness.

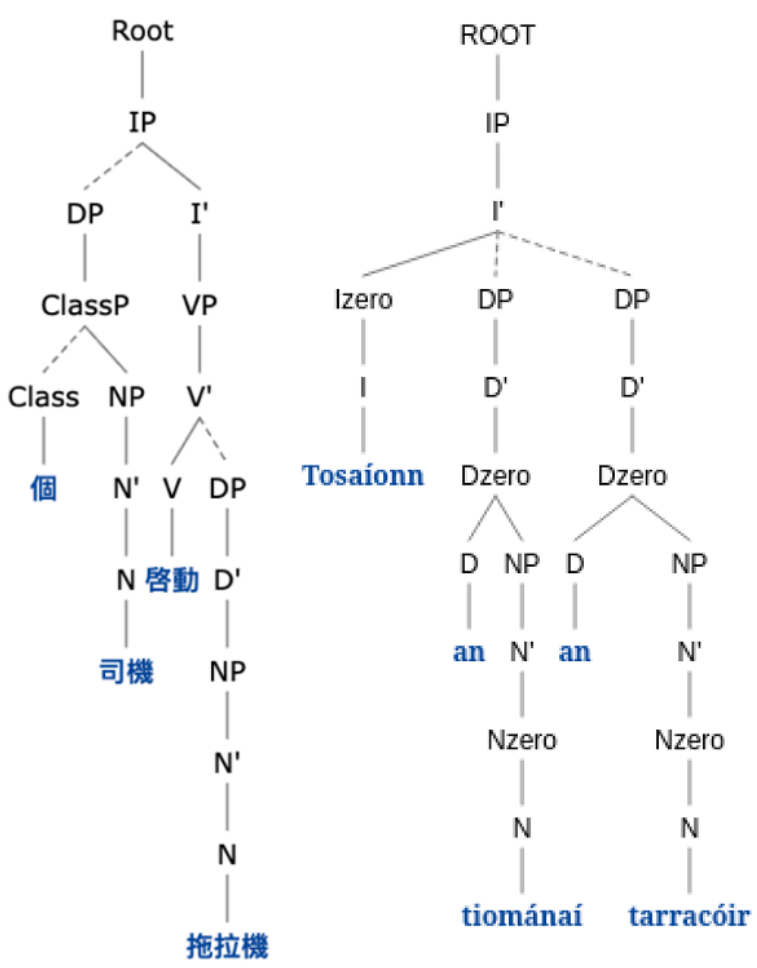

Figure 1: Cantonese and Irish c-structures for examples (1) and (2).

[3]The c- and f-structures presented in this paper were generated using the online INESS interface (Rosén et al., 2012), following the upload of the Cantonese and Irish LFG/XLE grammars to the server with assistance from Paul Meurer.

[4]An alternative Irish translation which may sound more natural to some speakers is *Tá dath dearg ar an tarracóir a cheannaigh an feirmeoir* 'There is a red color on the tractor that the farmer bought.'

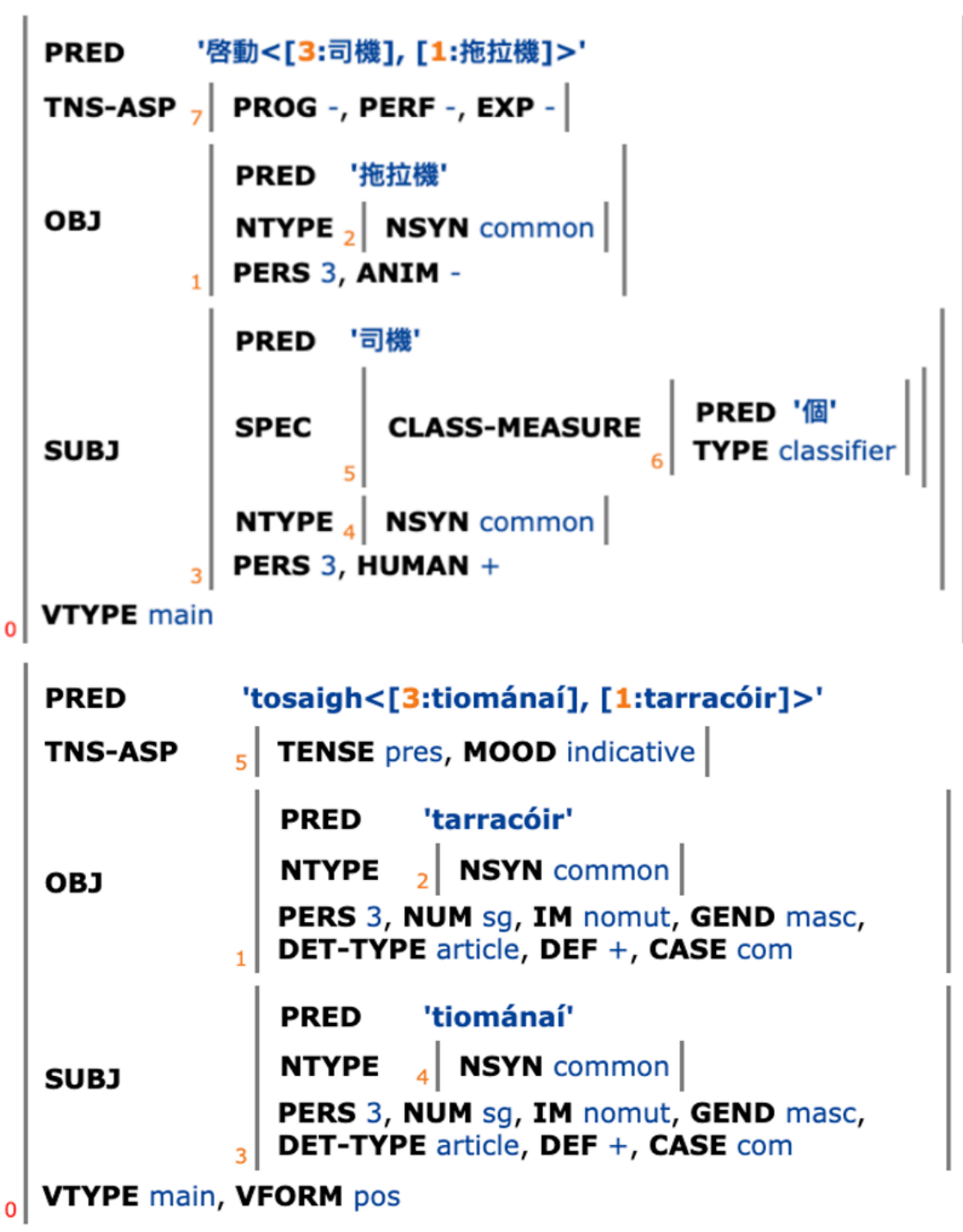

Figure 2: Cantonese and Irish f-structures for examples (1) and (2).

(3) 個 農夫 買 嘅 拖拉機 係 紅色-嘅
CL farmer buy REL tractor be red-ADJ
'The tractor that the farmer bought is red.' (Cantonese)

(4) Tá an tarracóir a cheannaigh an feirmeoir dearg
is the tractor that bought the farmer red
'The tractor that the farmer bought is red.' (Irish)

Another challenge concerns the translation of ParGramBank sentences. For example, in Irish, while the direct translation in (2) is grammatical, it may sound unnatural to some speakers because it does not capture the causative meaning of the sentence, namely *the driver causes the tractor to start*, rather than *the driver starts or begins an action themselves*. The alternative translation in (5) more accurately reflects this causative meaning.

Given that the ParGramBank family is constructed from translations of a parallel corpus, ensuring faithful and fluent translation is crucial. Poor translations may obscure underlying predicate–argument relations and complicate the establishment of cross-linguistic parallelism. In Section 3, we use these grammar-engineering tasks—translation and the construction of parallel syntactic representations—as a case study for investigating the methodological potential and limitations of LLM-assisted grammar engineering.

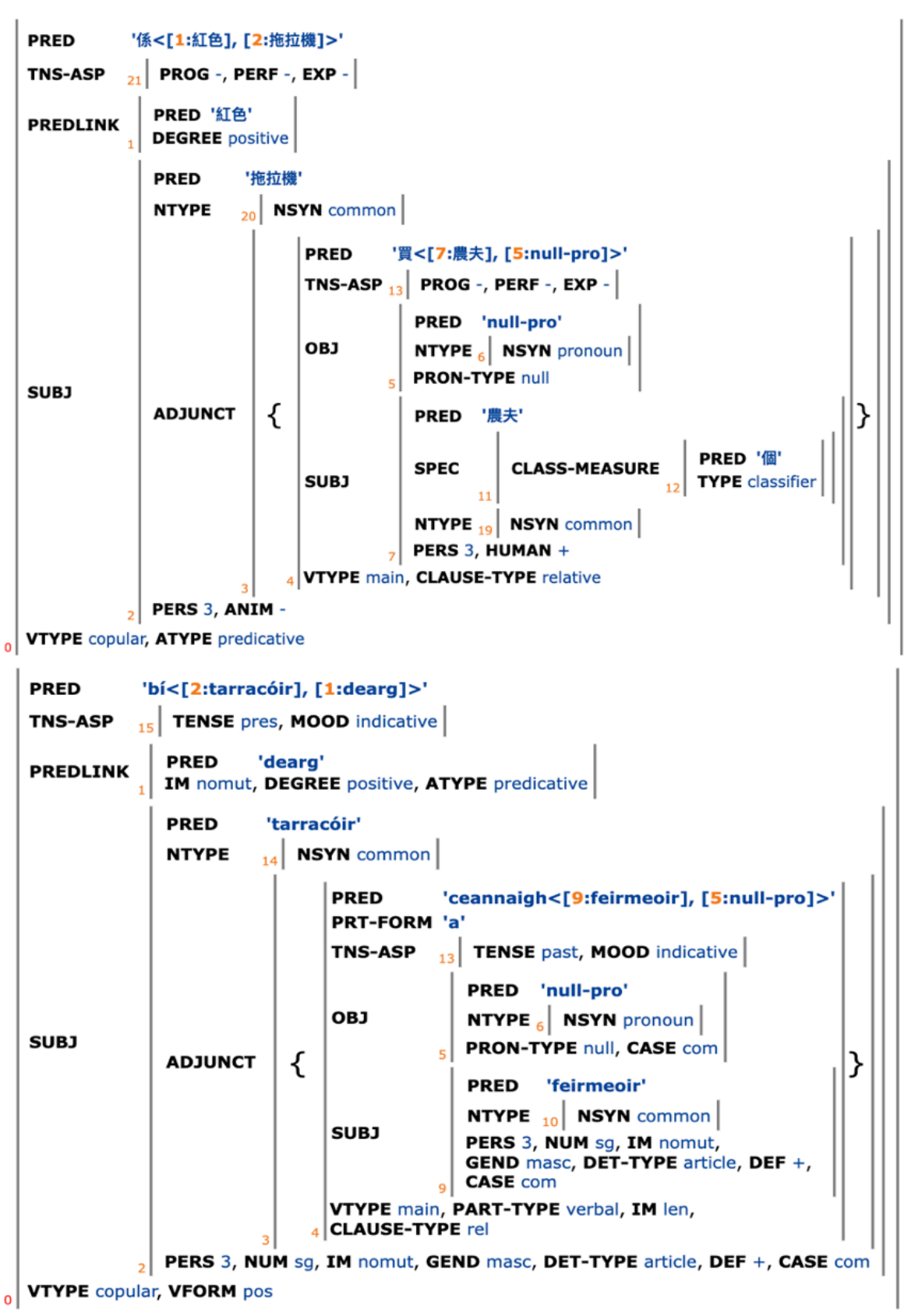

Figure 3: Cantonese and Irish f-structures for examples (3) and (4).

(5) Déanann an tiománaí an tarracóir a thosú
makes the driver the tractor to start
'The driver makes the tractor start.'

## 3 Probing LLMs: Cantonese–Irish translation and LFG analysis

### 3.1 Aim and objectives

In this section, we investigate the methodological potential and limitations of using multilingual LLMs to support grammar engineering tasks in three areas:

1. Producing f-structures that can serve as reference points for grammar development.
2. Capturing cross-linguistic similarities between Cantonese and Irish in f-structural formats to examine grammar parallelism.

3. Generating cross-linguistic translations between Cantonese and Irish to support the construction of parallel treebanks.

No previous work has investigated translation between Cantonese and Irish, both of which are under-resourced languages (McCrae and Doyle, 2019; Xiang et al., 2024). This study represents the first examination of this language pair.

## 3.2 Methodology

### 3.2.1 Tasks and prompts

We evaluated the `gpt-oss-120b` model (OpenAI) using a zero-shot prompting setup. It is one of two open-weight models released by OpenAI, alongside `gpt-oss-20b` (OpenAI et al., 2025), and has been tested on a range of NLP benchmarks (OpenAI et al., 2025; Bi et al., 2025). We selected `gpt-oss-120b` as a representative multilingual open-weight model and use it as a case study for investigating the methodological potential and limitations of LLM-assisted grammar engineering. The goal of this study is therefore to examine the kinds of support, errors, and limitations that may arise when a contemporary multilingual LLM is incorporated into a parallel grammar-engineering workflow. Inference was performed via Hugging Face's `InferenceClient` API with temperature set to zero. We conducted six evaluation tasks for Cantonese and Irish:

- Tasks 1 and 2: f-structure generation (English prompt)
- Tasks 3 and 4: translation combined with f-structure generation (English prompt)
- Tasks 5 and 6: translation only (Cantonese/Irish prompt)

In Task 1, the model generated an f-structure for each of the first 50 Cantonese sentences in ParGramBank, using the following prompt format:

```
You are an expert in Lexical Functional Grammar (LFG) and Cantonese. Provide one f-structure for the Cantonese sentence in triple single quotes '''...'''. Make sure the f-structure is in the attribute-value matrix format.
```

Task 2 followed the same procedure for Irish: the model was asked to generate an f-structure for each of the 50 Irish sentences. As a baseline, we evaluated the model on the first 50 English ParGramBank sentences (Butt et al., 1999; Sulger et al., 2013) using the same prompt format.

In Task 3, the model first translated each of the 50 Cantonese sentences into Irish and then generated an f-structure intended to reflect the shared structural properties of the source and translated sentences, using the following prompt format:

```
You are a linguist specializing in Lexical-Functional Grammar and a language expert in Cantonese and Irish. Your task is to first translate the Cantonese sentence in triple single quotes '''...''' into Irish; your translation should be both faithful and fluent. Then, provide one f-structure to capture the linguistic similarities between Cantonese and Irish with regard to this translated sentence.
```

Task 4 mirrored Task 3, but in the opposite direction: the model was asked to translate each of the 50 Irish sentences into Cantonese and then generate a corresponding f-structure.

Tasks 5 and 6 replicate the translation components of Tasks 3 and 4, respectively, but without f-structure generation: the model was asked to translate Cantonese sentences into Irish (Task 5) and Irish sentences into Cantonese (Task 6), with prompts formulated entirely in the respective source languages. These tasks were motivated by previous work (Cao et al., 2023), which suggests that varying the prompt language may affect model performance (cf. Hautli-Janisz et al., 2025).

### 3.2.2 Evaluation criteria

For tasks involving LLM-generated f-structures (Tasks 1–4), the Cantonese and Irish ParGram treebanks (Section 2) served as the gold standard for evaluation, while allowing for alternative linguistically valid analyses. We therefore adopted a four-point scale (**Excellent**, **Good**, **Fair**, **Bad**) to assess how well outputs conformed to LFG principles, prioritizing correct grammatical functions (GFs) and predicate–argument structure, with non-GF features (e.g., tense, aspect, person, number, and gender) considered secondarily. In brief, **Excellent** denotes a fully correct analysis, **Good** a structurally correct analysis with minor feature-level errors, **Fair** a partially correct analysis in which the core construction remains recognizable, and **Bad** an analysis that fails to capture the intended construction. Full details of the rating scheme are provided in Appendix B. F-structure evaluation was performed by the authors, who are Cantonese and Irish ParGram grammar engineers. The Cantonese and Irish ParGram treebanks served as the evaluation gold standards, and the treebanks have been discussed and refined through ongoing ParGram f-structure comparison meetings.[5] Evaluation of generated f-structures was performed entirely through expert manual assessment; no automated well-formedness checking or parser-based validation was applied.

For the translation tasks (Tasks 3–6), translation quality was evaluated independently in terms of faithfulness and fluency, as specified in the prompt, since these are not reliably captured by surface-form metrics such as BLEU. Given the absence of standard Cantonese–Irish reference translations, automatic evaluation is difficult to apply meaningfully. **Faithfulness** was defined as preservation of the core meaning of the source sentence, while **fluency** was assessed in terms of morphological and syntactic well-formedness as well as idiomaticity for native speakers of the target language. For each translation, faithfulness and fluency were judged separately using binary (Y/N) decisions. Each translation was evaluated independently by two annotators with expertise in the relevant languages. Disagreements were resolved through discussion. Inter-annotator agreement ranged from moderate to almost perfect (Cohen's $\kappa = 0.45$–$0.92$), as shown in Table 1. A translation was considered successful only if it was both faithful and fluent. We report faithfulness, fluency, and overall success rates separately.

| Traslation Task | Criterion | $\kappa$ |
|---|---|---|
| Irish→Cantonese (English prompt) | Faithfulness | 0.92 |
| Irish→Cantonese (English prompt) | Fluency | 0.62 |
| Irish→Cantonese (Irish prompt) | Faithfulness | 0.84 |
| Irish→Cantonese (Irish prompt) | Fluency | 0.71 |
| Cantonese→Irish (English prompt) | Faithfulness | 0.63 |
| Cantonese→Irish (English prompt) | Fluency | 0.75 |
| Cantonese→Irish (Cantonese prompt) | Faithfulness | 0.45 |
| Cantonese→Irish (Cantonese prompt) | Fluency | 0.63 |

Table 1: Inter-annotator agreement for translation evaluation, measured using Cohen's $\kappa$.

### 3.3 Results

The results of the translation tasks, based on the final adjudicated ratings, are presented in Table 2. We varied the prompt language with respect to both the source and target languages. For example, in the Cantonese-to-Irish tasks, prompts were provided in either English or Cantonese, simulating a scenario in which a Cantonese corpus is available and grammar engineers, who may not be familiar with Irish, translate Cantonese sentences into Irish to support the development of an Irish component. Conversely, for Irish-to-Cantonese translation, prompts were given in either English or Irish, reflecting a complementary scenario in which an Irish corpus is available and engineers translate Irish sentences into Cantonese to support the development of a Cantonese component.

For Irish-to-Cantonese translation, faithfulness scores (44–50%) were consistently higher than fluency scores (32–34%). Requiring translations to satisfy both criteria substantially reduced overall success rates (22–24%), suggesting that translations often preserved some aspects of meaning without producing fully natural target-language output.

The results for Cantonese-to-Irish translation showed the opposite trend, with faithfulness scores 8–12%, fluency scores 18–22%, and both criteria 6–8%. Higher fluency than faithfulness may reflect the lack of explicit tense and gender information in some of the Cantonese input, resulting in fluent but non-faithful Irish translations. Data sparsity issues associated with the more inflected language, and the models' propensity to invent words contributed to the lower faithfulness and fluency. However, if we ignore hallucinated words the fluency scores increase to 36-40%. The poorer overall results for Irish suggest that the LLM's training data may have contained less Irish data than Cantonese data.

For each source–target language pair, we conducted post hoc McNemar's tests to assess whether prompt language had a significant effect on overall translation success (i.e., translations judged both faithful and fluent). For Irish-to-Cantonese translation, an exact McNemar test on translations judged both faithful and fluent showed no significant difference between the English and Irish prompt conditions ($p = 1.00$). Likewise, for Cantonese-to-Irish translation, no significant difference was found between the English and Cantonese prompt conditions ($p = 1.00$). In other words, prompt language did not significantly affect translation performance for this language pair for our model. This finding contrasts with previous work such as Cao et al. (2023), but is consistent with Hautli-Janisz et al. (2025). Overall, trans-lation performance was unsatisfactory. We discuss recurring qualitative issues in Section 3.4.

[5] https://sites.google.com/view/pargram-meetings

| | Faithful | Fluent | Both |
|---|---|---|---|
| **Irish-to-Cantonese (English prompt)** | 50% | 32% | 22% |
| **Irish-to-Cantonese (Irish prompt)** | 44% | 34% | 24% |
| **Cantonese-to-Irish (English prompt)** | 12% | 18% | 6% |
| **Cantonese-to-Irish (Cantonese prompt)** | 8% | 22% | 8% |

Table 2: Percentages of translations judged faithful, fluent, and both faithful and fluent across translation tasks.

Table 3 presents the evaluation of LLM-generated f-structures for Tasks 1–4, including an English baseline. The LLM exhibited limited but non-negligible performance in generating LFG-style f-structures. The model performed best on English, with 20% of outputs rated as *Excellent* and 26% as *Good*, and a further 48% classified as *Fair*. For Cantonese (Task 1), 22% of outputs were rated as *Excellent* and 12% as *Good*, yielding 34% fully correct structural analyses. A further 40% were classified as *Fair*, indicating recognizable structures despite errors in GF assignment or subcategorization, while 26% were rated as *Bad*. For Irish (Task 2), only 2% of outputs were rated as *Excellent*, while 18% were *Good*, giving a total of 20% structurally correct analyses. As in Task 1, a substantial proportion (46%) fell into the *Fair* category, suggesting that the model often captured the general predicate–argument relations despite errors in GFs or feature specification; 34% were rated as *Bad*. Overall, while fully accurate analyses were relatively infrequent, the model captured at least some predicate–argument relations in around 70% of cases in both tasks.

To compare f-structure generation across English, Cantonese, and Irish, we coded the evaluation categories ordinally (*Excellent* = 4, *Good* = 3, *Fair* = 2, *Bad* = 1) and conducted a Friedman test, which showed a significant difference across conditions ($\chi^2(2) = 9.49, p = .009$). Pairwise Wilcoxon signed-rank tests indicated no significant difference between the English baseline and Cantonese (Task 1) ($p = .176$), but showed that both English ($Z = -3.84, p < .001, r = -.66$) and Cantonese ($Z = -2.16, p = .031, r = -.36$) significantly outperformed Irish (Task 2).

The lower ratings for Irish may be attributed to typological and data-related factors: its richer inflectional morphology requires more feature–value pairs in f-structure, increasing analytical complexity and making consistent specification more difficult, whereas Cantonese has minimal inflectional morphology, reducing this burden: once GFs are correctly identified, it is more likely for Cantonese analysis to receive the *Excellent* rating. The English baseline patterns more closely with Cantonese than with Irish, suggesting that morphologically simpler systems are handled more robustly by the model. In addition, the model may have been exposed to more Chinese than Irish data, although the extent to which this includes Cantonese is unclear.[6] Word order differences may also contribute: Cantonese and English are predominantly SVO, whereas Irish is VSO, a pattern that may be less well represented in the model's training data.

Performance on shared f-structure generation was weaker. In Task 3, 80% of sentences were rated as either *Fair* or *Bad*, with 44% classified as *Bad*. In Task 4, only 4% were rated as *Excellent*, while the majority fell into the *Fair* (50%) and *Bad* (38%) categories. A Wilcoxon signed-rank test over the four rating levels (*Excellent/Good/Fair/Bad*) showed that Task 1 received significantly higher ratings than Task 4 ($Z = -3.23, p = .001, r = -.65$). However, no statistically significant difference was found between Tasks 2 and 3 ($p = .991$). The additional requirement to abstract over cross-linguistic similarities appears to pose a greater challenge for the model.

### 3.4 Discussion: LLM-generated translation

Several recurring issues were observed in the LLM-generated translations. One prominent pattern is the intrusion of Mandarin forms in outputs intended to be Cantonese, affecting translation fluency. For example, in 我妹妹係一位好棒嘅老師 from Irish *Is múinteoir iontach í mo dheirfiúr* ('My sister is a great teacher'), the word 好棒 'great' is Mandarin rather than Cantonese; a more natural Cantonese equivalent would be 好叻. Similarly, in 佢哋為佢哋的女兒感到自豪 from Irish *Tá siad bródúil as a n-iníon* ('They are proud of their daughter'), the expression 的女兒 ('(someone's) daughter') is Mandarin, whereas Cantonese would typically use 個女 '(classifier) daughter'. Another example is the use of the Mandarin object-marking construction 把, where Cantonese would instead employ 將. These cases indicate that, despite being prompted for Cantonese output, the model often produced Mandarin lexical and grammatical forms, suggesting difficulty in maintaining a clear distinction between two Sinitic languages.

[6]See https://cameronrwolfe.substack.com/p/gpt-oss

| | Excellent | Good | Fair | Bad |
|---|---|---|---|---|
| **English f-structures (Baseline)** | 20% | 26% | 48% | 6% |
| **Cantonese f-structures (Task 1)** | 22% | 12% | 40% | 26% |
| **Irish f-structures (Task 2)** | 2% | 18% | 46% | 34% |
| **Cantonese–Irish shared f-structures (Task 3)** | 8% | 12% | 36% | 44% |
| **Irish–Cantonese shared f-structures (Task 4)** | 4% | 8% | 50% | 38% |

Table 3: Percentages of f-structures by evaluation category (Excellent, Good, Fair, Bad) in each task.

Another recurring issue concerns incorrect or distorted semantic interpretation in LLM-generated translations. For example, the Irish sentence *Rinne an buachaill é féin a fholcadh san abhainn* ('The boy bathed in the river') was translated into Cantonese as 那個男孩把自己淹死在河裡 ('The boy drowned himself in the river'). This output is not only in Mandarin but also introduces an incorrect causative 'drown' meaning instead of 'bathe'. Similarly, in translating *Tá an t-ádh ar cibé a cheannaigh an tarracóir seo* ('Whoever bought this tractor is a lucky person'), the model produced 呢個買家買咗嘅嘢都有好運 ('Whatever this buyer bought is lucky'), suggesting a misanalysis of the relative clause structure. In another case, *Chuir an feirmeoir iallach ar a mhac fágáil* ('The farmer made his son leave') was rendered as 農夫把小豬擺喺佢個仔身上 ('The farmer put a pig on his boy'), reflecting a failure to interpret the expression *Chuir ... iallach ar* ('to impose a constraint on/to force').

Invented words and incorrect translation of words were a common feature of the Irish translations. However, the level of inconsistency was surprising, and was clearly context-dependent. Of the 50 sentences in this corpus, 25 involve the word *tractor*. In 6 cases, the word *tarracóir* was correctly translated as 'tractor', but in 19 cases, it was variously translated as 'tarpaulin', 'driver', 'robber', 'burglar', 'rabbit', 'dealer', 'butcher', 'tarrock (a young gull)', 'translator', 'thief', and 'printer'. These predictions appear to be closely related to the context, e.g., 'printer' rather than 'tractor' in 'There is a problem with the ...' Other cases appear to be based on similarity to an English word, e.g., 'tarrock' for *tarracóir* , or an Irish word, e.g., *rósta* 'roasted' for *reoite* 'frozen'.

In the Cantonese-to-Irish translations, we observed a number of morphologically well-formed Irish neologisms. For example, the invented word *tráchtóir* was the most common translation for 'tractor', as well as *tróchar, trachtar, traicéad*, and *tráctor*. This phenomenon was previously reported for Irish by Castilho et al. (2025) and for Faroese by Scalvini et al. (2025).

### 3.5 Discussion: LLM-generated f-structures

In general, Irish sentences that mirror the predicate–argument structure of their English counterparts fare best, e.g., an active sentence with the ditransitive verb 'give' *Thug an feirmeoir seantarracóir dá chomharsa* ('The farmer gave an old tractor to his neighbour') is analyzed correctly, despite incorrect translations for *seantarracóir* as 'senior farmer' and *chomharsa* as 'companion'. However, many Irish sentences employ a construction different to English, e.g., *Lig an feirmeoir cnead as* 'The farmer let a groan out of him' instead of 'The farmer groaned'. This was plausibly but incorrectly guessed as 'The farmer let the herd out'.

Although the LLM did not always produce well-formed f-structures, it occasionally generated plausible alternative analyses that may be useful to grammar engineers. For example, in the Cantonese ParGram treebank, the prepositional phrase in (6) is analyzed as an ADJUNCT of 覺得 'feel', whereas the LLM treated it as an oblique argument. While this analysis is not adopted in the ParGram grammar, where 覺得 is consistently treated as a two-place predicate, it nonetheless suggests a possible alternative, namely allowing alternation between two-place and three-place structures.

(6) 佢哋 為 個 女 覺得 自豪
3PL for/about CL daughter feel proud
'They feel proud about their daughter.'

We observed that the LLM sometimes employed non-standard labels, such as REL for relative clauses instead of the LFG-preferred ADJUNCT. Although such labels fall outside the standard LFG feature inventory, they still suggest that the underlying structure is at least partially recognizable. LFG distinguishes between functions such as COMP, XCOMP, and PREDLINK (Dalrymple et al., 2019): PREDLINK marks predicative complements, COMP denotes closed complements (e.g., English *that*-clauses), and XCOMP denotes open complements typically found in control constructions (e.g., infinitival complements of *try*). While the LLM produced COMP and XCOMP, it sometimes misapplied these distinctions. In particular, relative clauses were occasionally analyzed as COMP, leading to *Bad* ratings and reflecting a conflation of modification with complementation—two fundamentally distinct notions in linguistic theory. The model also introduced thematic role labels such as AGENT into the f-structure, indicating further confusion between grammatical functions and thematic roles.

We observed potential instances of gender bias in the LLM-generated f-structures. For example, Cantonese nouns such as 農夫 'farmer' and 司機 'driver' were assigned masculine gender values, even though Cantonese does not mark grammatical gender. No contextual information in ParGramBank licenses such interpretations. For both Cantonese and Irish, the model appeared to assign gender features based on stereotypical world knowledge rather than linguistic evidence in the sentence. However, since the prompts did not explicitly specify the ParGram feature inventory, further investigation is needed to determine whether these assignments reflect genuine gender bias or alternative feature conventions adopted by the model.

## 4 LLM comparisons and ParGram treebanks

While this study focuses on a single multilingual LLM, it is important to situate the findings in relation to models with varying exposure to Cantonese and Irish. Some recent systems, e.g., EuroLLM (Ramos et al., 2026), NLLB (Team, 2024), and Irish-focused models such as UCCIX (Tran et al., 2024) and Qomhrá (McInerney et al., 2026), report greater incorporation of Irish data, while broadly multilingual models include varying amounts of Cantonese (Jiang et al., 2025). In contrast, large proprietary models (e.g., Claude (Enis and Hopkins, 2024)) are trained on substantially larger and more diverse datasets, which may influence performance on low-resource language pairs. Although a systematic comparison is beyond the scope of this paper, the usefulness of LLMs for grammar engineering likely depends on the quantity and quality of language-specific data, as well as task-specific adaptation. In this context, linguistically precise resources such as ParGram treebanks remain crucial: they provide gold-standard analyses for evaluation and benchmarks for assessing whether LLMs capture the formal generalizations required for grammar engineering.

Future work could explore cross-lingual transfer approaches that leverage high-resource languages such as English and Mandarin to support grammar engineering for lower-resource language pairs. Future work may also test models with more targeted language-specific training and fine-tuning, as well as varied prompting strategies and the use of LLMs in different collaborative grammar engineering settings.

## 5 Conclusions

Grammar engineering requires expertise in both linguistic formalism and computational implementation. This paper has presented the development of Cantonese and Irish ParGram grammars and treebanks, which maintain parallelism at an abstract functional level, alongside an evaluation of a multilingual LLM on translation and LFG-style f-structure generation. While the LLM produced some structurally meaningful outputs, overall performance was limited, particularly in shared f-structure tasks requiring cross-linguistic abstraction. Nonetheless, its outputs may still offer some reference value: even partially correct f-structures can capture aspects of predicate–argument relations and suggest alternative analyses. However, such outputs require careful manual verification and cannot replace expert-driven grammar development.

## Acknowledgments

The first author would like to acknowledge the support of a Research Project Grant from the Leverhulme Trust for the project *Modelling Coreference Resolution across Syntax, Semantics, and Discourse* (RPG-2025-064). Both authors are grateful to Mary Dalrymple for her feedback during the preparation of this manuscript. We also thank members of the ParGram f-structure comparison meetings for their continuous feedback and support since 2023. We are grateful to our language consultants, Gearóid Ó Donnchadha and Lun Kin Lo, for their assistance with the translation rating tasks. Finally, we thank the three anonymous reviewers for their helpful comments and suggestions, and the BriGap program committee for their consideration of this work.

## A ParGramBank Sentences

**English ParGram Sentences (1–50):**
These sentences are openly accessible at: https://clarino.uib.no/iness/home > Treebanks > English and ParGram > View and search the selected treebanks.

1. The driver starts the tractor.
2. The tractor is red.
3. What did the farmer see?
4. Did the farmer sell his tractor?
5. Push the button.
6. Don't push the button.
7. The farmer gave his neighbor an old tractor.
8. The farmer cut the tree down.
9. The farmer groaned.
10. My neighbor was given an old tractor by the farmer.
11. The tree was cut down yesterday.
12. The tree had been cut down.

13. The tractor starts with a shudder.
14. The tractor appeared.
15. The boy knows the tractor is red.
16. The child thinks he started the tractor.
17. The farmer knows who started the tractor.
18. The child wondered whether the button had been pushed.
19. The tractor that the farmer bought is red.
20. The man who bought the tractor left.
21. The store the farmer bought the tractor from closed.
22. Whoever bought this tractor is a lucky person.
23. The farmer made his son clean the tractor.
24. The farmer made his son leave.
25. The farmer made her son buy the tractor.
26. The farmer let her son buy the tractor.
27. The farmer bought his son a tractor.
28. The woman bought the tractor for her husband.
29. The lovers danced until dawn.
30. The boy bathed in the river.
31. The teacher read to himself aloud.
32. The brothers bought the tractor for each other.
33. My sister is a great teacher.
34. The child is in the house.
35. The children are happy.
36. It is raining.
37. There is a problem with the tractor.
38. The book cover depicted a tractor.
39. Let's get ice-cream.
40. The boy swept up the broken wine bottle.
41. The red wine bottle broke.
42. There are great green globs of greasy grimy gopher guts.
43. They are proud of their daughter.
44. My tractor is faster than your sports car.
45. My tractor is the fastest vehicle in the county.
46. The barking dog woke the neighbours.
47. The tea drinking woman admired her new purchase from eBay.
48. The farmer wants to buy a tractor.
49. The farmer's daughter promised to repair the tractor.
50. The farmer persuaded his wife to buy a new tractor.

**Cantonese ParGram Sentences (1–50):**
The following sentences are the idiomatic Cantonese translation of the respective ParGram sentences.

1. 個司機啓動拖拉機
2. 架拖拉機係紅色嘅
3. 個農夫睇到乜嘢
4. 個農夫賣咗佢嘅拖拉機呀
5. 撳個制
6. 唔好撳個制
7. 個農夫送畀佢嘅隔離屋一架舊拖拉機
8. 個農夫將棵樹斬落嚟
9. 個農夫嘆氣
10. 個農夫送咗一架舊拖拉機畀我隔離屋
11. 棵樹琴日俾人斬咗落嚟
12. 棵樹俾人斬咗落嚟
13. 個拖拉機顫咗一陣之後啓動咗
14. 架拖拉機出現咗
15. 個男仔知道架拖拉機係紅色嘅
16. 個細路仔以為佢啓動咗架拖拉機
17. 個農夫知道邊個啓動咗架拖拉機

18. 個細路仔想知個制係咪俾人撳咗落去
19. 個農夫買嘅拖拉機係紅色嘅
20. 買嗰架拖拉機嘅男人走咗
21. 個農夫買拖拉機嘅舖頭收咗檔
22. 買咗呢架拖拉機嘅人好幸運
23. 個農夫叫佢個仔清理拖拉機
24. 個農夫叫佢個仔走
25. 個農夫叫佢個仔買咗一架拖拉機
26. 個農夫俾佢個仔買咗一架拖拉機
27. 個農夫幫佢個仔買咗一架拖拉機
28. 個女人買咗呢架拖拉機畀佢老公
29. 呢對情侶一直跳舞到天光
30. 個男仔喺條河度沖涼
31. 個老師大聲咁讀畀自己聽
32. 兄弟互相買咗架拖拉機
33. 我家姐係一位好老師
34. 嗰個細路喺屋入面
35. 啲細路好開心
36. 而家落緊雨
37. 架拖拉機有問題
38. 書嘅封面畫咗一架拖拉機
39. 我哋去食雪糕啦
40. 個男仔將酒樽嘅碎片掃起嚟
41. 個紅酒樽爆咗
42. 有一大團油淋淋又污糟嘅地鼠内臟
43. 佢哋為個女覺得自豪
44. 我架拖拉機比你架跑車快
45. 我架拖拉機係成個縣最快嘅交通工具
46. 隻吠緊嘅狗嘈醒咗啲鄰居
47. 個飲緊茶嘅女人欣賞佢喺 eBay 買返嚟嘅新嘢
48. 個農夫想買一架拖拉機
49. 個農夫嘅女應承修理架拖拉機
50. 個農夫勸佢老婆買一架新拖拉機

**Irish ParGram Sentences (1–50):**
The following sentences are the idiomatic Irish translation of the respective ParGram sentences.

1. Tosaíonn an tiománaí an tarracóir.
2. Tá dath dearg ar an tarracóir.
3. Céard a chonaic an feirmeoir?
4. Ar dhíol an feirmeoir a tharracóir?
5. Brúigh an cnaipe.
6. Ná brúigh an cnaipe.
7. Thug an feirmeoir seantarracóir dá chomharsa.
8. Ghearr an feirmeoir an crann anuas.
9. Lig an feirmeoir cnead as.
10. Bhí seantarracóir tugtha do mo chomharsa ag an bhfeirmeoir.
11. Gearradh an crann anuas inné.
12. Bhí an crann gearrtha anuas.
13. Tosaíonn an tarracóir le creathán.
14. Nocht an tarracóir.
15. Tá a fhios ag an mbuachaill go bhfuil dath dearg ar an tarracóir.
16. Ceapann an páiste gur chuir sé an tarracóir ar siúl.
17. Tá a fhios ag an bhfeirmeoir cé a chuir an tarracóir ar siúl.
18. Bhí an páiste ag déanamh iontais de ar bhrúdh an cnaipe nó nár bhrúdh.
19. Tá dath dearg ar an tarracóir a cheannaigh an feirmeoir.
20. D’fhág an fear a cheannaigh an tarracóir.
21. Dúnadh an siopa ónar cheannaigh an feirmeoir an tarracóir.
22. Tá an t-ádh ar cibé a cheannaigh an tarracóir seo.

23. Chuir an feirmeoir iachall ar a mhac an tarracóir a ghlanadh.
24. Chuir an feirmeoir iachall ar a mhac fágáil.
25. Chuir an feirmeoir iachall ar a mac an tarracóir a cheannach.
26. Lig an feirmeoir dá mac an tarracóir a cheannach.
27. Cheannaigh an feirmeoir tarracóir dá mhac.
28. Cheannaigh an bhean an tarracóir dá fear céile.
29. Bhí na leannán ag damhsa go breacadh an lae.
30. Rinne an buachaill é féin a fholcadh san abhainn.
31. Léigh an múinteoir dó féin os ard .
32. Cheannaigh na deartháireacha an tarracóir dá chéile.
33. Is múinteoir iontach í mo dheirfiúr.
34. Tá an páiste sa teach.
35. Tá na páistí sásta.
36. Tá sé ag cur báistí.
37. Tá fadhb leis an tarracóir.
38. Léirigh clúdach an leabhair tarracóir.
39. Faighimis uachtar reoite.
40. Scuab an buachaill suas an buidéal fíona briste.
41. Bhris an buidéal fíona dheirg.
42. Tá blobaí móra glasa de phutóga gófair gréisceacha brocacha ann.
43. Tá siad bródúil as a n-iníon.
44. Tá mo tharracóir níos tapúla ná do charr spóirt.
45. Is é mo tharracóir an fheiticil is tapúla sa chontae.
46. Dhúisigh tafann an mhadra na comharsana.
47. D'fhéach bean ólta an tae le sásamh ar a ceannachán úr ó eBay.
48. Ba mhaith leis an bhfeirmeoir tarracóir a cheannach.
49. Gheall iníon an fheirmeora go ndeiseodh sí an tarracóir.
50. Chuir an feirmeoir ina luí ar a bhean chéile tarracóir nua a cheannach.

## B Rating Scheme for F-structure Evaluation

- Excellent (Fully correct)
  - All grammatical functions (GFs) are correctly specified (e.g. SUBJ, OBJ, OBL, COMP, XCOMP, ADJUNCT)
  - Subcategorization frame <...> is correct
  - All relevant non-GF features (e.g., TENSE, ASPECT, NUMBER, GENDER, CASE, PERSON) are correct
- Good (Structurally correct, some incorrect features)
  - All GFs are correct and consistent with the predicate argument structure
  - Subcategorization frame is correct
  - Only issues in non-GF features
- Fair (Partially correct analysis)
  - The core construction is recognizable, but there is at least one GF error, such as a missing argument or incorrect GF assignment (e.g. OBJ vs OBL) or a mismatch with the subcategorization frame
  - May include errors in non-GF features
- Bad (Not recognizable construction)
  - The core construction is not recognizable due to major errors in GFs
  - The f-structure is incompatible with the intended predicate-argument structure; e.g., relative clause or adjunct analyzed as complement